\pdfoutput=1
\documentclass[11pt]{article}

\usepackage[]{naacl2021}

\usepackage{times}
\usepackage{latexsym}
\usepackage{amsmath}
\DeclareMathOperator*{\argmax}{arg\,max}

\usepackage{graphicx}

\usepackage[]{xcolor}
\usepackage{tabularx}
\usepackage{multirow}
\usepackage{paralist}

\usepackage[T1]{fontenc}
\usepackage[utf8]{inputenc}

\usepackage{microtype}

\newcommand{\PaperName}{TSLM }

\title{\vspace*{-0.5in}
{{\small \hfill Published in NAACL'2021}\\
\vspace*{.25in}} Time-Stamped Language Model: Teaching Language Models to Understand the Flow of Events}

\author{Hossein Rajaby Faghihi \\
  Michigan State University \\
  \texttt{rajabyfa@msu.edu} \\\And
  Parisa Kordjamshidi \\
  Michigan State University \\
  \texttt{kordjams@msu.edu} \\}

\begin{document}
\maketitle
\begin{abstract}
% \headline{what do we do to enhance LM}
Tracking entities throughout a procedure described in a text is challenging due to the dynamic nature of the world described in the process. Firstly, we propose to formulate this task as a question answering problem. This enables us to use pre-trained transformer-based language models on other QA benchmarks by adapting those to the procedural text understanding. Secondly, since the transformer-based language models cannot encode the flow of events by themselves, we propose a Time-Stamped Language Model~(\PaperName model) to encode event information in LMs architecture by introducing the timestamp encoding. Our model evaluated on the Propara dataset shows improvements on the published state-of-the-art results with a $3.1\%$ increase in F1 score. Moreover, our model yields better results on the location prediction task on the NPN-Cooking dataset. This result indicates that our approach is effective for procedural text understanding in general.
\end{abstract}

\section{Introduction}
% \headline{What is procedural reasoning}
A procedural text such as a recipe or an instruction usually describes the interaction between multiple entities and their attribute changes at each step of a process. For example, the photosynthesis procedure can contain steps such as 1. \textit{Roots absorb water from soil}; 2. \textit{The water flows to the leaf}; 3. \textit{Light from the sun and CO2 enter the leaf}; 4. \textit{The water, light, and CO2 combine into a mixture}; 5. \textit{Mixture forms sugar}.
Procedural text understanding is a machine reading comprehension task defined on procedural texts. Answering questions such as "what is the location of the mixture at step 4", in the above example, requires tracking entities' interactions to predict their attributes at each step~\cite{mishra2018tracking,bosselut2017simulating}. This is quite challenging due to the dynamic nature of the entities' attributes in the context.

Transformer-based language models have shown promising results on multi-hop or single-hop question answering benchmarks such as HotpotQA~\cite{yang2018hotpotqa}, SQuAD~\cite{rajpurkar2016squad}, and Drop~\cite{dua2019drop}. However, it is hard to expect LMs to understand the flow of events and pay attention to the time in the procedure~(e.g., step 4) without extra modeling efforts.

In recent research, different approaches are taken to address procedural reasoning based on language models using QA formulations. Following the intuition that attributes of entities can be retrieved based on the current and previous steps,  DynaPro~\cite{amini2020procedural} modifies the input to only contain those sentences in the input at each time. This will provide a different input to the model based on each question to help it detect changes after adding each step. KG-MRC~\cite{das2018building} also generates a dynamic knowledge graph at each step to answer the questions. However, this intuition is contradicted in some scenarios such as detecting inputs of the process. For instance, the answer to the question "Where is light as step 0?" is "Sun", even if it is not mentioned in the first sentence of the process. Inputs are entities that are not created in the process.

The architecture of the QA transformer-based LMs is very similar to the traditional attention mechanism. Other methods such as ProLocal~\cite{mishra2018tracking} and ProGlobal~\cite{mishra2018tracking} have structured this task by finding the attention of each entity to the text at each step. To be sensitive to the changes at each step, ProLocal manually changes the model's input by removing all steps except the one related to the question. ProGlobal computes attention to the whole context while adding a distance value. Distance value is computed for each token based on its distance to the direct mention of the entity at each step.

The current language models convey rich linguistic knowledge and can serve as a strong basis for solving various NLP tasks~\cite{liu2019roberta,devlin-etal-2019-bert,yang2019xlnet}. That is why most of the state-of-the-art models on procedural reasoning are also built based on current language models~\cite{amini2020procedural,gupta2019tracking}. Following the same idea, we investigate the challenges that current models are facing for dealing with procedural text and propose a new approach for feeding the procedural information into LMs in a way that the LM-based QA models are aware of the taken steps and can answer the questions related to each specific step in the procedure.

We propose the Time-Stamped Language model~(\PaperName model), which uses timestamp embedding to encode past, current, and future time of events as a part of the input to the model. \PaperName utilizes timestamp embedding to answer differently to the same question and context based on different steps of the process.
As we do not change the portion of the input manually, our approach enables us to benefit from the pre-trained LMs on other QA benchmarks by using their parameters to initialize our model and adapt their architecture by introducing a new embedding type. Here, we use RoBERTa~\cite{liu2019roberta} as our baseline language model.

We evaluate our model on two benchmarks, Propara~\cite{mishra2018tracking} and NPN-Cooking~\cite{bosselut2017simulating}. Propara contains procedural paragraphs describing a series of events with detailed annotations of the entities along with their status and location. NPN-Cooking contains cooking recipes annotated with their ingredients and their changes after each step in criteria such as location, cleanliness, and temperature.

\PaperName differs from previous research as its primary focus is on using pre-trained QA models and integrating the flow of events in the global representation of the text rather than manually changing the part of the input fed to the model at each step. \PaperName outperforms the state-of-the-art models in nearly all metrics of two different evaluations defined on the Propara dataset. Results show a $3.1\%$ F1 score improvement and a $10.4\%$ improvement in recall. \PaperName also achieves the state-of-the-art result on the location accuracy on the NPN-Cooking location change prediction task by a margin of $1.55\%$. 
% Since our proposal does not change the base architecture of a transformer model, other architectural components of common-sense reasoning methods can be easily integrated with it.
In summary, our contribution is as follows:
\begin{compactitem}
    \item We propose Time-Stamped Language Model~(\PaperName model) to encode the meaning of past, present, and future steps in processing a procedural text in language models.
    \item Our proposal enables procedural text understanding models to benefit from pre-trained LM-based QA models on general-domain QA benchmarks.
    \item \PaperName outperforms the state-of-the-art models on the Propara benchmark on both document-level and sentence-level evaluations. \PaperName improves the performance state-of-the-art models on the location prediction task of the NPN-Cooking~\cite{bosselut2017simulating} benchmark. 
    \item Improving over two different procedural text understanding benchmarks suggests that our approach is effective, in general, for solving the problems that require  the integration of the flow of events in a process.
\end{compactitem}

\begin{table*}
    \centering
    \begin{tabular}{|l|c|c|c|c|c|c|}
    \hline
         &&\multicolumn{5}{c|}{Participants} \\ \hline
         Paragraph& State number& Water & Light & CO2 & Mixture & Sugar \\ \hline
         (Before the process starts) & State 0 & Soil & Sun & ? & - & - \\ \hline
          \small{Roots absorb water from soil} & State 1 & Root & Sun & ? & - & - \\ \hline
          \small{The water flows to the leaf} & State 2 & Leaf & Sun & ? & - & - \\ \hline
          \small{Light from the sun and CO2 enter the leaf} & State 3 & Leaf & Leaf & Leaf & - & - \\ \hline
          \small{The water, light, and CO2 combine into a mixture} & State 4 & - & - & - & Leaf & - \\ \hline
          \small{Mixture forms sugar} & State 5 & - & - & - & - & Leaf \\ \hline
         
    \end{tabular}
    \caption{An example of procedural text and its annotations from the Propara dataset~\cite{mishra2018tracking}. "-" means entity does not exist. "?" means the location of entity is unknown.}
    \label{table:sample_text}
\end{table*}

\section{Problem Definition}
An example of a procedural text is shown in Table \ref{table:sample_text}. The example is taken from the Propara~\cite{mishra2018tracking} dataset and shows the photosynthesis procedure. At each row, the first column is list of the sentences, each of which forms one step of the procedure. The second column contains the number of the step in the process and the rest are the entities interacting in the process and their location at each step. The location of entities at step 0 is their initial location, which is not affected by this process. If an entity has a known or unknown location~(specified by ``?'') at step 0, we call it an input.

The procedural text understanding task is defined as follows. Given a procedure p containing a list of $n$ sentences  $P=\{s_1,...s_n\}$, an entity $e$ and a time step $t_i$, we find $L$, the location of that entity and specify the status $S$ of that entity. Status S is one value in a predefined set $S=\{\text{non-existence, unknown-location, known-location}\}$. location $L$ is a span of text in the procedure that is specified with its beginning and end token.
We formulate the task as finding function $F$ that maps each triplet of entity, procedure and time step to a pair of entity location and status: $(S,L) = F(e,P,t_{i})$

% \pk{I rephrased it above if that is ok refine the text and formula symbols and you can remove the below two paragraphs then.}

% We formulate the procedural text understanding task as $S,L = F(e,P,t_{i})$, \pk{I don't understand this structure: "we formulate the task as and then showing a formula." } where $e$ is an entity in the process, $P$ is a procedure described as a text, and $t_{i}$ is a step in the procedure. $S$ is the status and $L$ is the location of entity $e$ at step $t_{i}$ of the process $P$. 

% Candidate answers for $S$ are "Non-existence", Unknown Location", and  "Known location". ``Non-existence'' indicates that the entity does not exist, ``Unknown Location'' means that the entity exists and its location is not known, and the ``Known location'' means that the entity exists, and its location is known. $L$ is also computed as a span of the text in $P$, where $P$ is a list of sentences $[s_{1}, s_{2}, ..., s_{n}]$ and $n$ is the number of steps in the process.

\section{Proposed Procedural Reasoning Model}
\subsection{QA Setting}
To predict the status and the location of entities at each step, we model $F$ with a question answering setting. For each entity $e$, we form the input $Q_e$ as follows:
\begin{equation}
    \begin{aligned}
        Q_{e} = &\text{[CLS] Where is $e$? [SEP]} \\
        &\text{$s_{1}$ [SEP] $s_{2}$ [SEP] ..., $s_{n}$ [SEP]}
    \end{aligned}
\end{equation}
% We always ask the question about the location of the entity in the form of "Where is" question type.

Although $Q_{e}$ is not a step-dependent representation and does not incorporate any different information for each step, our mapping function needs to generate different answers for the question "Where is entity $e$?" based on each step of the procedure.
For instance, consider the example in Table \ref{table:sample_text} and the question "where is water?", our model should generate different answers at four different steps. 
The answer will be ``root'', ``leaf'', ``leaf'', ``non-existence'' for steps 1 to 4, respectively.
%~( $\xrightarrow{\text{answers}}$1:root, 2:leaf, 3:none-existence, 4:none-existence)\pk{the last piece is not readable}\comment{Is the updated one better now?}

To model this, we create pairs of $(Q_{e}, t_{i})$ for each $i \in \{0, 1, ..., n\}$. For each pair, $Q_{e}$ is timestamped according to $t_{i}$ using $Timestamp(.)$ function described in Sec. \ref{sec:timestamp} and mapped to an updated step-dependent representation, $Q_{e}^{t_{i}} = Timestamp(Q_{e}, t_i)$.

The updated input representation is fed to a language model~(here ROBERTA) to obtain the step-dependent entity representation, $R_{e}^{t_{i}}$, as shown in Equation \ref{formula:roberta}. We discuss the special case of $i = 0$ in more details in Sec. \ref{sec:timestamp}.

\begin{equation}
\label{formula:roberta}
    R_{e}^{t_{i}} = RoBERTa(Q_{e}^{t_{i}})
\end{equation}

% \pk{you can bring the formula with ROBERTA to here Re=...}

%\pk{our proposed model: this is vague, you still do not address the issue by feeding xe into it, maybe say we feed xe and the time embedding as it is shown in the figure and then continue describing the figure} 

% \headline{Describe how the QA model overview is and what are the predictions}
\noindent We use the step-dependent entity representation, $R_{e}^{t_{i}}$, and forward it to another mapping function $g(.)$ to obtain the location and status of the entity $e$ in the output. In particular the output includes the following three vectors, a vector representing the predictions of entity status $S$, another vector for each token's probability of being the start of the location span $L$, and a third vector carrying the probability of each word being the last token of the location span.
The outputs of the model are computed according to the Equation \ref{formula:output}.
\begin{equation}
\label{formula:output}
        (status, Start\_prob, End\_prob) = g(R_{e}^{t_{i}})
\end{equation}
where $R_{e}$ is the tokens' representations output of RoBERTa~\cite{liu2019roberta}, and $g(.)$ is a function we apply on the token representations to get the final predictions. We will discuss each part of the model separately in the following sections.

\begin{figure*}[ht]
    \centering
    \includegraphics[width=\linewidth]{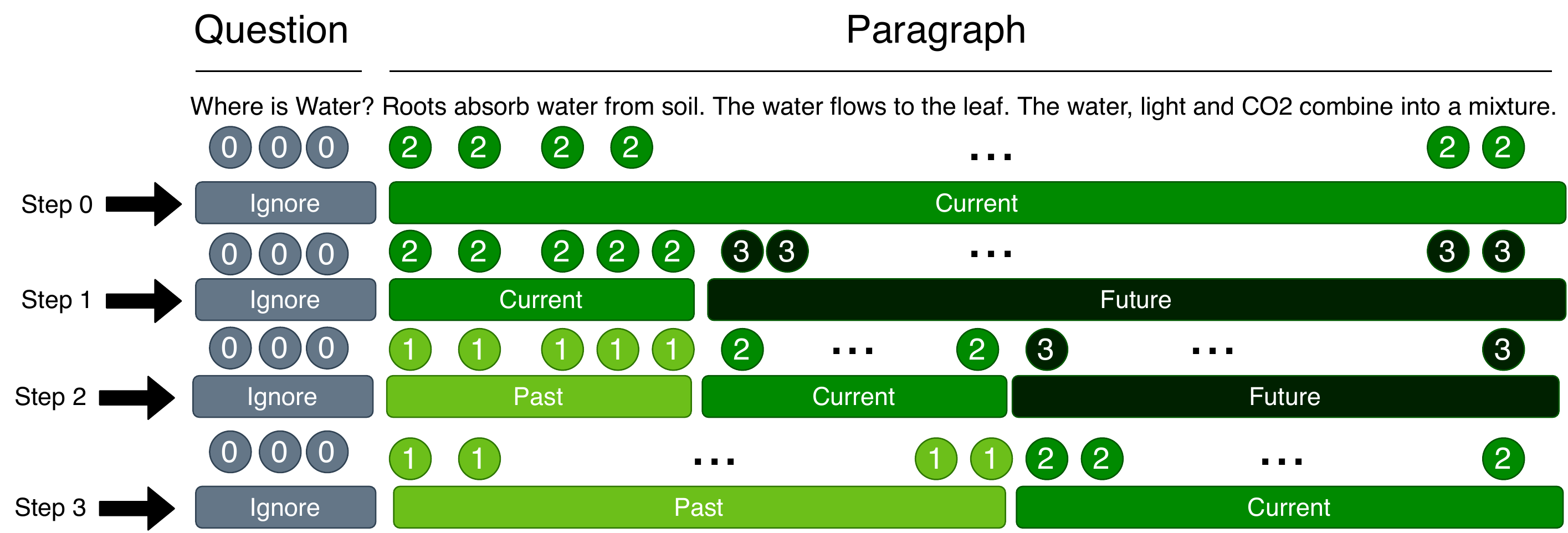}
    \caption{An example of timestamp embedding in a procedural text. The question is always ignored with value "0". At each step $i$, the tokens from that step are paired with ``current'' value, tokens from steps 0 to $i$ are paired with ``past'' value, and the tokens from step $i$ to last step are paired with the ``future'' value.}
    \label{fig:timestamp}
\end{figure*}

\subsection{Timestamp Embedding}
\label{sec:timestamp}
%\pk{the $t_i$ should be the timestamp embedding in the above formulas?OK }\comment{$t_{i}$ is just a number in upper setting}
% \headline{Describe the addition of the timestamp embedding}
% \pk{why you do not continue write formulas for this module? this is not consistent, you have partly formulas and partly refer to image boxes, you need both, it should not be details but your final output should some out of all your formulas at the end}\comment{That may be not easily understood, I though the image is more clear, let me then change the above formulas before calling Roberta and add a function for timestamp embedding} \pk{OK you can just say Attention().. no need for details of that}\comment{I have updated previous and below passage based on this.}
The timestamp embedding adds the step information to the input $Q_{e}$ to be considered in the attention mechanism. The step attention is designed to distinguish between current~(what is happening now), past~(what has happened before), and future~(what has not yet happened) information.

\begin{figure}[]
    \centering
    \includegraphics[width=\linewidth]{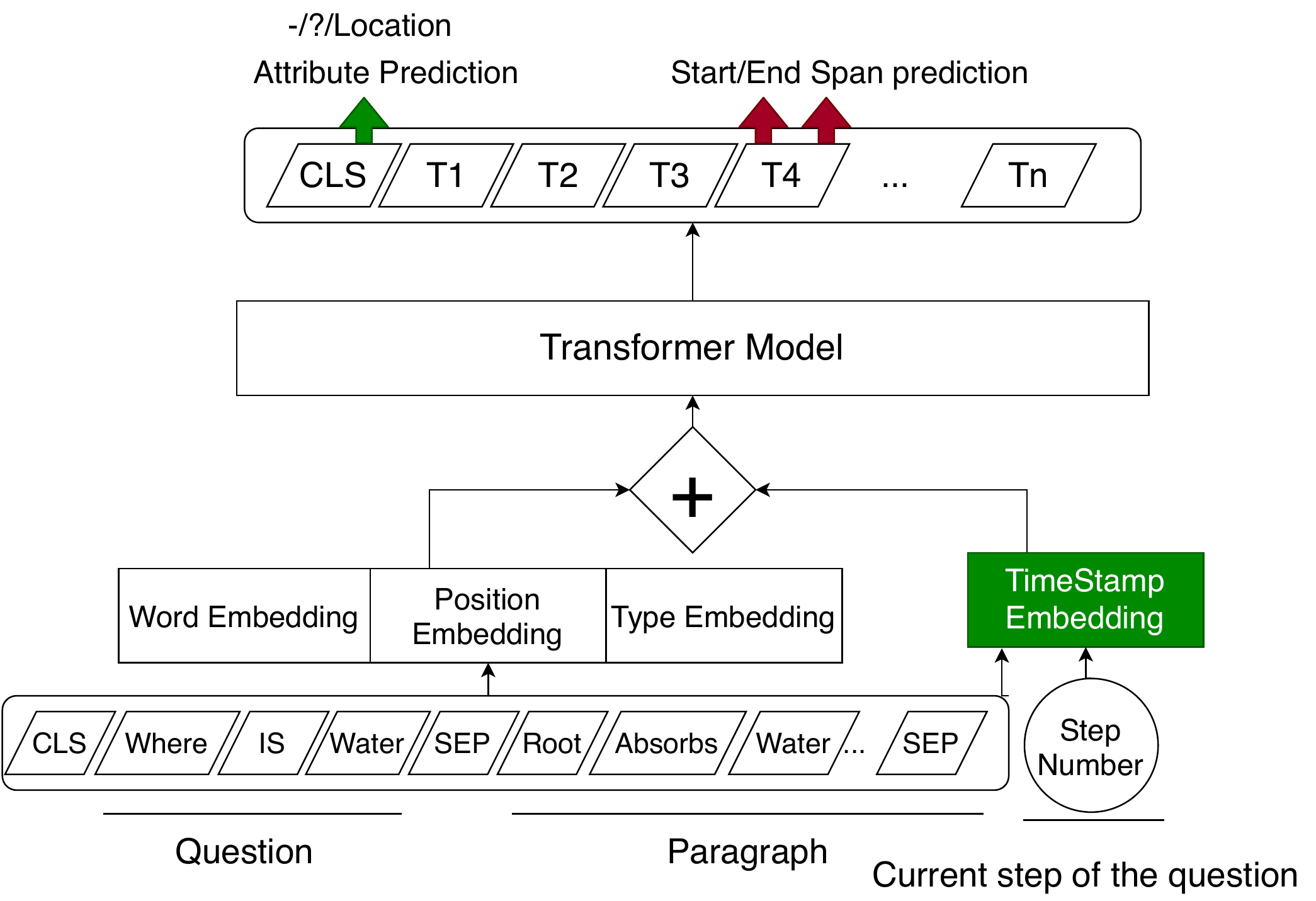}
    \caption{An overview of the proposed model. The ``Timestamp Embedding'' module is introduced in this work and the rest of the modules are taken from basic language model architecture.}
    \label{fig:model}
\end{figure}

We use the mapping function $Timestamp(.)$ from the pair $(Q_{e}, t_{i})$ to add a number along with each token in $Q_{e}$ and retrieve the step-dependent input $Q_{e}^{t_{i}}$ as shown in Figure \ref{fig:timestamp}.
The Mapping function $Timestamp(.)$ integrates past, current, and future representations to all of the tokens related to each part. $Timestamp(.)$ function assigns number $1$ for past, $2$ for current, and $3$ for future tokens in the paragraph by considering one step of the process as the current event. These values are used to compute an embedding vector for each token, which will be added to its initial representation as shown in Figure \ref{fig:model}. The special number $0$ is assigned to the question tokens, which are not part of the process timeline. For predicting State 0~(The inputs of the process), we set all the paragraph information as the current step. 
% \pk{not sure why the question should be ignored? isn't that important? just from the step it is not clear which entity you are talking about? I may miss something here. What do you mean by ignore maybe is the right question. if this is because you only care about timestamp embedding? then point to it clearly, and mention that where the question information is considered}
% \headline{Give one example of the timestamp embedding with Past, Present, Future}

\subsection{Status classification}
% \headline{Describe the classifier of None/?/Location}
To predict the entities' status, we apply a linear classification module on top of the $[CLS]$ token representation in $R_{e}$ as shown in Equation \ref{formula:attribute}. 
\begin{equation}
\label{formula:attribute}
    Attribute = Softmax(W^T Re_{[C]})
\end{equation}
where $Re_{[C]}$ is the representation of the $[CLS]$ token which is the first token in $R_{e}$.

\subsection{Span prediction}
% \headline{Describe the start/end selection from the tokens}
% \headline{Describe the start/end selection from the tokens}
%\pk{I say "for each" because to my understanding you run the model for each question and a single step separately, when you say "at each" it gives me the impression that your model does all steps together but sequentially one at a time, is that what you do?}\comment{I do all steps at a time in a batch for one entity, but I guess in this setting I am just talking about one entity and one step so I can use "for" if that is more accurate}. \pk{the main question is that what is exactly one training example? to my understanding each example includes only one timestamp and one entity. If you form the batch in a specific way that is a separate thing, also the number of steps probably are different for each procedure so I guess you can not put everything related to one entity in one batch always, right?} \comment{Entities are defined for each procedure, not globally. But yes, in this setting it is each entity each step so I will change to for}

We predict a location span for each entity for each step of the process as shown in Equation \ref{formula:span}, we follow the popular approach of selecting start/end tokens to detect a span of the text as the final answer. We compute the probability of each token being the start or the end of the answer span. If the index with the highest probability to be the start token is $token_{start}$ and for the end token is $token_{end}$, the answer location will be $Location = P[token_{start}:token_{end}]$.

\begin{equation}
\label{formula:span}
    \begin{aligned}
        \text{Start\_prob} &= \textrm{Softmax}(W_{\text{start}}^T R_{e}^{t_{i}}) \\
        \text{End\_prob} &= \textrm{Softmax}(W_{\text{end}}^T R_{e}^{t_{i}}) \\
        \text{token}_{\text{start}} &= \argmax_i(\text{Start\_prob}) \\
        \text{token}_{\text{end}} &= \argmax_i(\text{End\_prob})
    \end{aligned}
\end{equation}
\subsection{Training}
% \headline{Describe the joint training and the batch-wise operation on entity and steps}
We use the cross-entropy loss function to train the model. At each prediction for entity $e$ at timestamp $t_{i}$, we compute one loss value $loss_{attribute}$ regarding the status prediction and one loss value $loss_{location}$ for the span selection. The variable $loss_{location}$ is the summation of the losses of the start token and the end token prediction, $loss_{location} = loss_{location_{start}} + loss_{location_{end}}$. The final loss of entity $e$ at time $t_{i}$ is computed as in Equation \ref{formula:loss}.
\begin{equation}
\label{formula:loss}
    Loss^{e}_{i} = loss_{(i,attribute)}^{e} + loss_{(i,location)}^{e}
\end{equation}
% \pk{end of the equation is end of the sentence, so we need to use "."}
\subsection{Inference}
% \headline{Describe how we change the output of the span prediction based on pre-defined selections based on Pos tagging}
At inference time, we apply two different post-processing rules on the outputs of the model. First, we impose that the final selected location answer should be a noun phrase in the original procedure. Considering that a location span is a noun phrase, we limit the model to do a $softmax$ over tokens of noun phrases in the paragraph to select the start and end tokens. Second, we apply consistency rules to make sure that our predicted status of entities are consistent. We define the two following rules: 
\begin{itemize}
    \item An entity can not be created if it has been already destroyed : \textit{if $S_e^{t_{i}}$ is "non-existence" and $S_e^{t_{i+1}}$ is unknown or known location, then for every step $j$, if $S_e^{t_{j}}$ is unknown or known location and $S_e^{t_{j+1}}$ is "non-existence", then $i$ < $j$}.
    
    \item An entity cannot be created/destroyed twice in a process: \textit{if $S_e^{t_{j}}$ and $S_e^{t_{i}}$ are both "-", $S_e^{t_{j+1}}$ and $S^{t_{i+1}}$ are both either known or unknown location, then $i = j$}.
\end{itemize}
$S_e^{t_{i}}$ is the status of entity $e$ at step $t_{i}$ of the process.
% ~(Going from "-" at steps $i$ and $j$ to "?" or a location at steps $i+1$ and $j+1$, where $i \neq j$). 

We do not apply an optimization/search algorithm to find the best assignment over the predictions according to the defined constraints. The constraints are only applied based on the order of the steps to ensure that the later predictions are consistent with the ones made before. 
%\pk{? is this the right word, what do you mean by "left to right assumption"}.\comment{I revised it, is this better?pk:OK}

\section{Experiments}
\subsection{Datasets}
% \headline{Describe the Dataset}

\textbf{Propara~\cite{mishra2018tracking}:} This dataset was created as a benchmark for procedural text understanding to track entities at each step of a process. Propara contains 488 paragraphs and 3,300 sentences with annotations that are provided by crowd-workers. The annotations~(~81,000) are the location of entities at each step of the process. The location can be either the name of the location, unknown location, or specified as non-existence. 

\noindent \textbf{NPN-Cooking~\cite{bosselut2017simulating}:} This is a benchmark containing textual cooking instructions. Annotators have specified ingredients of these recipes and explained the recipe using different changes happening on each ingredient at each step of the instructions. These changes are reported in categories such as location, temperature, cleanliness, and shape. We evaluate our model on the location prediction task of this benchmark, which is the hardest task due to having more than 260 candidate answers. We do not use the candidates to find the locations in our setting; Instead, we find a span of the text as the final location answer. This is a relatively harder setting but more flexible and generalizable than the classification setting. 

\subsection{Implementation Details}
% \headline{Describe the parameter setup}
We use SGD optimizer implemented by Pytorch~\cite{paszke2017automatic} to update the model parameters. The learning rate for the Propara implementation is set to $3-e4$ and is updated by a scheduler with a $0.5$ coefficient every 50 steps. We use $1-e6$ as the learning rate and a scheduler with $0.5$ coefficient to update the parameters every ten steps on the NPN-Cooking implementation. The implementation code is publicly available at GitHub\footnote{https://github.com/HLR/TSLM}.

% \headline{Describe the batch and implementation framework}
We use RoBERTa~\cite{liu2019roberta} question answering architecture provided by HuggingFace~\cite{Wolf2019HuggingFacesTS}. RoBERTa is pretrained with SQuAD~\cite{rajpurkar2016squad} and used as our base language model to compute the token representations. Our model executes batches containing an entity at every step and makes updates based on the average loss of entities per procedure. The network parameters are updated after executing one whole example. The implementation code will be publicly available on GitHub after acceptance. 
\subsection{Evaluation}
\label{sec:evaluation}
% \headline{List the results of all models}
\textbf{Sentence-level} evaluation is introduced in \cite{mishra2018tracking} for Propara dataset. This evaluation focuses on the following three categories.
\begin{compactitem}
    \item \textbf{Cat1} Is $e$ created~(destroyed/moved) during the process?
    \item \textbf{Cat2} When is $e$ created~(destroyed/moved) during the process?
    \item \textbf{Cat3} Where is $e$ created~(destroyed/moved from or to) during the process?
\end{compactitem}
\begin{table}
    \centering
    \begin{tabular}{|c|c|c|c|c|}
    \hline
         Step & Entity & Action & Before & After\\ \hline
         1& Water& Move& Root& Leaf \\
         2& Water&Destroy &Leaf & - \\
         1& Sugar&Create& - & Leaf \\
         2 &Sugar & None & Leaf & Leaf \\ \hline
    \end{tabular}
    \caption{A sample table to evaluate the Propara document-level task.}
    \label{table:evaluation_sample}
\end{table}

\begin{table*}
    % \centering
    \begin{tabular}{l|ccccc|ccc}
         &\multicolumn{5}{c|}{Sentence-level} & \multicolumn{3}{c}{Document-level} \\ \hline
         Model& Cat1 & Cat2 & Cat3 & Macro-Avg & Micro-Avg & P & R & F1 \\ \hline
         ProLocal~\cite{mishra2018tracking} &62.7& 30.5& 10.4& 34.5& 34.0& \textbf{77.4}& 22.9& 35.3 \\
         ProGlobal~\cite{mishra2018tracking} &63.0& 36.4& 35.9 &45.1& 45.4& 46.7& 52.4& 49.4\\
         EntNet~\cite{henaff2016tracking} &51.6& 18.8& 7.8& 26.1& 26.0& 50.2& 33.5& 40.2 \\
         QRN~\cite{seo2016query} &52.4& 15.5 &10.9 &26.3 &26.5 &55.5& 31.3 &40.0 \\
         KG-MRC~\cite{das2018building} &62.9& 40.0& 38.2& 47.0& 46.6& 64.5& 50.7& 56.8\\
         NCET~\cite{gupta2019tracking} &73.7 &47.1 &41.0 &53.9 &54.0 &67.1 &58.5 &62.5 \\
         XPAD~\cite{dalvi2019everything} &- &- &- &-& - &70.5& 45.3& 55.2 \\
         ProStruct~\cite{tandon2018reasoning}& - &- &-& -& -& 74.3& 43.0& 54.5 \\ 
         DYNAPRO~\cite{amini2020procedural} &72.4 &49.3 &\textbf{44.5} &55.4 &55.5 &75.2 &58.0 &65.5 \\ \hline
         \PaperName~(Our Model) & \textbf{78.81}& \textbf{56.798}& 40.9& \textbf{58.83} &\textbf{58.37} & 68.4& \textbf{68.9}&\textbf{68.6}\\ \hline
    \end{tabular}
    \caption{Results from sentence-level and document-level evaluation On Propara. C${i}$ evaluations are defined Section \ref{sec:evaluation}.}
    \label{table:results}
\end{table*}
\noindent \textbf{Document-level} evaluation is a more comprehensive evaluation process and introduced later in \cite{tandon2018reasoning} for Propara benchmark. Currently, this is the default evaluation in the Propara leaderboard containing four criteria: 
\begin{compactitem}
    \item \textbf{What are the Inputs?} Which entities existed before the process began and do not exist after the process ends.
    \item \textbf{What are the Outputs?} Which entities got created during the process?
    \item \textbf{What are the Conversions?} Which entities got converted to other entities?
    \item \textbf{What are the Moves? } Which entities moved from one location to another?
\end{compactitem} 
The document-level evaluation requires models to reformat their predictions in a tabular format as shown in Table \ref{table:evaluation_sample}. At each row of this table, for each entity at a specific step, we can see the action applied on that entity, the location of that entity before that step, and the location of the entity after that step.
Action takes values from a predefined set including, ``None'', ``Create'', ``Move'', and ``Destroy''. The exact action can be specified based on the before and after locations.
% \pk{maybe give a small example one case that shows this is computable}\comment{I have the following , should I move up? : The action column is filled based on the data provided in before and after locations. If the before location is/is not "-" and after location is not/is "-", then the action is "Create"/"Destroy". If the before and after locations are equal, then the action is "None" and if the before and after locations are both spans and are different from each other, the action is "Move".}\pk{Oh ok no need to move}

We have to process our (Status $S$, Location $L$) predictions at each step to generate a similar tabular format as in Table \ref{table:evaluation_sample}. We define $r^{i}_e$ as a row in this table which stores the predictions related to entity $e$ at step $t_{i}$. To fill this row, we first process the status predictions. If the status prediction $S$ is either ``-'' or ``?'', we fill those values directly in the after location column. The before location column value of $r^{i}_e$ is always equal to the after location column value of $r^{i-1}_e$. If the status is predicted to be a ``Known Location'', we fill the predicted location span $L$ into the after location column of $r^{i}_e$.  

The action column is filled based on the data provided in before and after locations columns. If the before location is/isn't "-" and after location is not/is "-", then the action is "Create"/"Destroy". If the before and after locations are equal, then the action is "None" and if the before and after locations are both spans and are different from each other, the action is "Move".

\noindent \textbf{NPN-Cooking location change: }
We evaluate our model on the NPN-Cooking benchmark by computing the accuracy of the predicted locations at steps where the locations of ingredients change. We use the portion of the data that has been annotated by the location changes to train and evaluate our model. In this evaluation, we do not use the status prediction part of our proposed \PaperName model. Since training our model on the whole training set takes a very long time~(around 20 hours per iteration), we use a reduced number of samples for training. This is a practice that is also used in other prior work~\cite{das2018building}. 

\subsection{Results}
\begin{table*}[]
    \centering
    \begin{tabular}{l|c|c|l}
         Model& Accuracy & Training Samples & Prediction task \\ \hline
         NPN-cooking~\cite{bosselut2017simulating}& 51.3 & \small{$\sim83,000$~(all data)} & Classification \\
         KG-MRC~\cite{das2018building} & 51.6 & \small{$\sim10,000$} & Span Prediction \\
         DynaPro~\cite{amini2020procedural} & 62.9 & \small{$\sim83,000$~(all data)} & Classification \\
         \hline
         \multirow{2}{*}{\PaperName~(Our Model)} & 63.73 & \small{$\sim10,000$} & Span Prediction \\
         & \textbf{64.45} & \small{$\sim15,000$} & Span Prediction \\
    \end{tabular}
    \caption{Results on the NPN-Cooking benchmark. Both class prediction and span prediction tasks are the same but use two different settings, one selects among candidates, and the other chooses a span from the recipe. However, each model has used a different setting and a different portion of the training data. The information of the data splits was not available that makes a fair comparison hard.}
    \label{table:npn-cooking}
\end{table*}

The performance of our model on Propara dataset~\cite{mishra2018tracking} is quantified in Table \ref{table:results}. 
% \headline{Describe advantage of our model in F1 and Recall comparing to previous approaches}
Results show that our model improves the SOTA by a $3.1\%$ margin in the F1 score and improves the Recall metric with $10.4\%$ on the document-level evaluation. On the sentence-level evaluation, we outperform SOTA models with a $5.11\%$ in Cat1, and $7.49\%$ in Cat2 and by a $~3.4\%$ margin in the macro-average. We report Table \ref{table:results} without considering the consistency rules and evaluate the effect of those in the ablation study in Sec. \ref{sec:ablation}.
% \pk{you mean these are not your best results?}\comment{The best result is when we have no consistency rule applied}\pk{so is it interesting to talk about it at all? because technically that is just and additional post-processing}\comment{Actually I added this in general because every other paper also has some kind of approach to makes consistency, but do you mean to remove it from this sentence or the whole constraints in general? I have also mentioned in the future direction that we want to replace these rules with optimization algorithms} \pk{if this is common for this task then keep it, OK}

In Table \ref{table:detail}, we report a more detailed quantified analysis of \PaperName model's performance based on each different criteria defined in the document-level evaluation. Table \ref{table:detail} shows that our model performs best on detecting the procedure's outputs and performs worst on detecting the moves. Detecting moves is essentially hard for \PaperName as it is predicting outputs based on the whole paragraph at once. Outperforming SOTA results on the input and output detection suggests that \PaperName model can understand the interactions between entities and detect the entities which exist before the process begins. The detection of input entities is one of the weak aspects of the previous research that we improve here.
\begin{table}[]
    \centering
    \begin{tabular}{l|c|c|c}
         Criteria& Precision & Recall & F1  \\ \hline
         Inputs& 89.8& 71.3& 79.5\\
         Outputs& 85.6&91.4&88.4 \\
         Conversions&57.7 &56.7 & 57.2\\
         Moves& 40.5&56&47 \\
    \end{tabular}
    \caption{Detailed analysis of \PaperName performance on the Propara test set on four criteria defined in the document-level evaluation.}
    \label{table:detail}
\end{table}

A recent unpublished research~\cite{zhang2020knowledge} reports better results than our model. However, their primary focus is on common-sense reasoning and their goal is orthogonal to our main focus in proposing \PaperName model. Such approaches can be later integrated with \PaperName to benefit from common-sense knowledge on solving the Propara dataset. 

The reason that \PaperName performs better at \textit{recall} and worse at \textit{precision} is that our model looks at the global context, which increases the recall and lowers the precision when local information is strongly important. The same phenomenon (better recall) is observed in ProGlobal, which also considers global information as we do, compared to ProLocal.

Table~\ref{table:npn-cooking} shows our results on the NPN-Cooking benchmark for the location prediction task. Results are computed by only considering the steps that contain a location change and are reported by computing the accuracy of predicting those changes. Our results show that \PaperName outperforms the SOTA models with a $1.55\%$ margin on accuracy even after training on 15,000 training samples. To be comparable with the KG-MRC~\cite{das2018building} experiment on NPN-Cooking which is only trained on 10k samples, we report the performance of our model trained on the same number of samples, where \PaperName gets a $12.1\%$ improvement over the performance of KG-MRC~\cite{das2018building}.

\subsection{Ablation Study}
\label{sec:ablation}
To evaluate the importance of each module one at a time, we report the performance of the \PaperName by removing the noun-phrase filtering at inference, the consistency rules, timestamp embedding, SQuAD~\cite{rajpurkar2016squad} pre-training, and by replacing RoBERTa~\cite{liu2019roberta} with BERT~\cite{devlin-etal-2019-bert}. These variations are evaluated on the development set of the Propara dataset and reported in Table \ref{table:ablation}. As stated before and shown in Table~\ref{table:ablation}, it is impossible to remove the timestamp embedding as that is the only part of the model enabling changes in the answer at each step. Hence, by removing that, the model cannot converge and yields a $~25\%$ decrease on the F1 score. The simple consistency and span filtering rules are relatively easy to be learned by the model based on the available data, therefore adding those does not affect the final performance of the model. 

TSLM$_\text{BERT}$ experiment is designed to ensure a fair comparison with previous research~\cite{amini2020procedural} which has used BERT as their base language model. The comparison of \textit{TSLM$_{BERT}$} to \textit{-SQuAD Pre-training} and \textit{- Timestamp Embedding} in Table \ref{table:ablation} indicates that using RoBERTa instead of BERT is not as much important as our main proposal~(using Time-stamp encoding) in \PaperName model. Also, TSLM$_\text{BERT}$ achieves $66.7\%$ F1 score on the Propara test set, which is $1.2\%$ better than the current SOTA performance.

By removing the SQuAD pre-training phase, the model performance drops with a $10.6\%$ in the F1 score. This indicates that despite the difference between the procedural text understanding and the general MRC tasks, it is quite beneficial to design methods that can transfer knowledge from other QA data sources to help with procedural reasoning. This is crucial as annotating procedural texts is relatively more expensive and time-consuming.
\begin{table}[h]
    % \centering
    \begin{tabular}{l|ccc}
    \hline
         Model&P & R & F1  \\ \hline
         TSLM$_\text{RoBERTa}$  & 72.9 & 74.1 & \textbf{73.5} \\ \hline
         - constraints& 73.8 & 73.3 & \textbf{73.5} \\
         - noun-phrase filtering &73.5 & 73.3 & 73.4\\ 
         - SQuAD Pre-training& 78.8&52.2& 62.8\\ 
         - Timestamp Embedding &94.6&32.6&48.5 \\ \hline
         TSLM$_\text{BERT}$ &69.2& 73.5&71.3 \\ \hline
    \end{tabular}
    \caption{Ablation study results on the development set of the Propara document-level task. ``- constraints'', ``- Span filtering'', and ``- Timestamp Encoding'' shows our model performance while removing those modules. \textit{-SQuAD Pre-training} is when we do not pre-train our base language model on SQuAD. TSLM$_\text{BERT}$ is when we use BERT as the base language model.}
    \label{table:ablation}
\end{table}

\section{Discussion}
We provide more samples to support our hypothesis in solving the procedural reasoning task and answer some of the main questions about the ideas presented in \PaperName model.

\noindent\textbf{Why is the whole context important?}
The main intuition behind \PaperName is that the whole context, not just previous information, matters in reasoning over a process. Here, we provide some samples from Propara to show why this intuition is correct. 
Consider this partial paragraph, "Step $i$: With enough time the pressure builds up greatly. Step $i+1$: The resulting volcano may explode.". Looking at the annotated status and location, the "volcano" is being created at Step $i$ without even being mentioned in that step. This is only detectable if we look at the next step saying "The resulting Volcano...". 

As another example, consider this partial paragraph: "Step $i$: Dead plants form layers called peat. ... Step $i+3$: Pressure squeezes water out of the peat.". 
The annotation indicates that the location of "water" is being changed to "peat" at step $i$, which is only possible to detect if the model is aware of the following steps indicating that the water comes out of the peat. 

\noindent\textbf{Positional Embedding VS Time-stamp encoding}:
As mentioned before the whole context (future and past events) is essential for procedural reasoning at a specific step. However, the reasoning should focus on one step at a time, given the whole context. While positional encoding encodes the order of information at the token-level for reasoning over the entire text, we need another level of encoding to specify the steps' positions (boundaries) and, more importantly, to indicate the step that the model should focus on when answering a question.

\noindent\textbf{Advantages/Disadvantages of \PaperName model}:
\PaperName integrates higher-level information into the token representations. This higher-level information can come from event-sequence~(time of events), sentence-level, or any other higher source than the token-level information. The first advantage of \PaperName is that it enables designing a model which is aware of the whole context, while previous methods had to customize the input at each step to only contain the information of earlier steps. Furthermore, using \PaperName enables us to use pretrained QA models on other datasets without requiring us to retrain them with the added time-stamped encoding. One main disadvantage of \PaperName model, which is natural due to the larger context setting in this model, is not being sensitive to local changes, which is consistent with the observation in the comparison between ProGlobal and ProLocal models.

\section{Related Works}
% MRC\footnote{Machine Reading Comprehension} is an interesting task investigated with many recent benchmarks like SQuAD~\cite{rajpurkar2016squad}, HotPotQA~\cite{yang2018hotpotqa}, and Drop~\cite{dua2019drop}. Language models such as BERT~\cite{devlin-etal-2019-bert}, RoBERTa~\cite{liu2019roberta}, and XLNET~\cite{yang2019xlnet} have shown very promising results on such benchmarks. However, as the setting in procedural text understanding is more challenging due to the dynamic nature of the described phenomena in the text, it calls for special procedural benchmarks and procedure-aware models.

ScoNe~\cite{long2016simpler}, NPN-Cooking~\cite{bosselut2017simulating}, bAbI~\cite{weston2015towards}, ProcessBank~\cite{berant2014modeling}, and Propara~\cite{mishra2018tracking} are benchmarks proposed to evaluate models on procedural text understanding. Processbank~\cite{berant2014modeling} contains procedural paragraphs mainly concentrated on extracting arguments and relations for the events rather than tracking the states of entities. ScoNe~\cite{long2016simpler} aims to handle co-reference in a procedural text expressed about a simulated environment. bAbI~\cite{weston2015towards} is a simpler machine-generated textual dataset containing multiple procedural tasks such as motion tracking, which has encouraged the community to develop neural network models supporting explicit modeling of memories~\cite{weston2014memory,santoro2018relational} and gated recurrent models~\cite{cho2014properties,henaff2016tracking}. NPN-Cooking~\cite{bosselut2017simulating} contains recipes annotated with the state changes of ingredients on criteria such as location, temperature, and composition. Propara~\cite{mishra2018tracking} provides procedural paragraphs and detailed annotations of entity locations and the status of their existence at each step of a process.

% Some earlier models on procedural text understanding, inspired by bAbI dataset, are Memory Network architecture~\cite{weston2014memory}, Recurrent Relational network~(RRN)~\cite{santoro2018relational}, and gated recurrent models such as GRU~\cite{cho2014properties} and Recurrent
% Entity Networks(EntNet)~\cite{henaff2016tracking}. EntNet uses dynamic memory for the world's hidden state, which will be updated based on a gated mechanism at each step. RRN augments neural networks with the capacity to do multi-step relational reasoning while keeping a memory of hidden states~(separate memory per entity) at each step and updating that based on the memory representations and newly retrieved representations at each step. 

Inspired by Propara and NPN-Cooking benchmarks, recent research has focused on tracking entities in a procedural text. Query Reduction Networks~(QRN)~\cite{seo2016query} performs gated propagation of a hidden state vector at each step. Neural Process Network (NPN)~\cite{bosselut2017simulating} computes the state changes at each step by looking at the predicted actions and involved entities.
Prolocal~\cite{mishra2018tracking} predicts locations and status changes locally based on each sentence and then globally propagates the predictions using a persistence rule. Proglobal~\cite{mishra2018tracking} predicts the status changes and locations over the whole paragraph using distance values at each step and predicts current status based on current representation and the predictions of the previous step. ProStruct~\cite{tandon2018reasoning} aims to integrate manually extracted rules or knowledge-base information on VerbNet~\cite{schuler2005verbnet} as constraints to inject common-sense into the model. KG-MRC~\cite{das2018building} uses a dynamic knowledge graph of entities over time and predicts locations with spans of the text by utilizing reading comprehension models. Ncet~\cite{gupta2019tracking} updates entities representation based on each sentence and connects sentences together with an LSTM. To ensure the consistency of predictions, Ncet uses a neural CRF over the changing entity representations. XPAD~\cite{dalvi2019everything} is also proposed to make dependency graphs on the Propara dataset to explain the dependencies of events over time. Most recently, DynaPro~\cite{amini2020procedural} feeds an incremental input to pre-trained LMs' question answering architecture to predict entity status and transitions jointly. 

\PaperName differs from recent research, as we propose a simple, straightforward, and effective technique to make our model benefit from pre-trained LMs on general MRC tasks and yet enhance their ability to operate on procedural text understanding. We explicitly inject past, current, and future timestamps into the language models input and implicitly train the model to understand the events' flow rather than manually feeding different portions of the context at each step. Procedural reasoning has also been pursued within the multi-modality domain~\cite{yagcioglu2018recipeqa,faghihi2020latent,amac2019procedural} which has additional challenges of aligning the representation spaces of different modalities.

\section{Conclusion}
% \headline{Summary of the model}
We proposed the Time-Stamped Language Model~(\PaperName model), a novel approach based on a simple and effective idea, which enables pre-trained QA models to process procedural texts and produce different outputs based on each step to track entities and their changes. \PaperName utilizes a timestamp function that causes the attention modules in the transformer-based LM architecture to incorporate past, current, and future information by computing a timestamp embedding for each input token. 
% input side of the models to \pk{help in computing: compute} a timestamp embedding for each token in its own context \pk{usually instead of parenthesis you can use "i.e., Past, Current, Future.}"~(Past, Current, Future).
% \headline{Summary of the results}
Our experiments show a $3.1\%$ improvement on the F1 score and a $10.4\%$ improvement over the Recall metric on Propara Dataset. Our model further outperforms the state-of-the-art models with a $1.55\%$ margin in the NPN-Cooking dataset accuracy for the location prediction task.

% \headline{Future works}
As a future direction, it is worth investigating how common-sense knowledge can be integrated with the \PaperName setting by augmenting the process context using external sources of related domain knowledge. We also intend to investigate the effectiveness of our approach on similar tasks on other domains and benchmarks. As another future direction, it can be effective to apply an inference algorithm to impose the global consistency constraints over joint predictions in procedural reasoning instead of using naive post-processing rules.
% \section*{Acknowledgements}
\section*{Acknowledgements}

This project is partially funded by National Science Foundation~(NSF) CAREER Award \texttt{\#}2028626 and the Office of Naval Research~(ONR) grant \texttt{\#}N00014-20-1-2005. 

\bibliography{anthology,custom}

\begin{thebibliography}{28}
\expandafter\ifx\csname natexlab\endcsname\relax\def\natexlab#1{#1}\fi

\bibitem[{Amac et~al.(2019)Amac, Yagcioglu, Erdem, and
  Erdem}]{amac2019procedural}
Mustafa~Sercan Amac, Semih Yagcioglu, Aykut Erdem, and Erkut Erdem. 2019.
\newblock Procedural reasoning networks for understanding multimodal
  procedures.
\newblock In \emph{Proceedings of the 23rd Conference on Computational Natural
  Language Learning (CoNLL)}, pages 441--451.

\bibitem[{Amini et~al.(2020)Amini, Bosselut, Mishra, Choi, and
  Hajishirzi}]{amini2020procedural}
Aida Amini, Antoine Bosselut, Bhavana~Dalvi Mishra, Yejin Choi, and Hannaneh
  Hajishirzi. 2020.
\newblock Procedural reading comprehension with attribute-aware context flow.
\newblock In \emph{Proceedings of the Conference on Automated Knowledge Base
  Construction (AKBC)}.

\bibitem[{Berant et~al.(2014)Berant, Srikumar, Chen, Vander~Linden, Harding,
  Huang, Clark, and Manning}]{berant2014modeling}
Jonathan Berant, Vivek Srikumar, Pei-Chun Chen, Abby Vander~Linden, Brittany
  Harding, Brad Huang, Peter Clark, and Christopher~D Manning. 2014.
\newblock Modeling biological processes for reading comprehension.
\newblock In \emph{Proceedings of the 2014 Conference on Empirical Methods in
  Natural Language Processing (EMNLP)}, pages 1499--1510.

\bibitem[{Bosselut et~al.(2018)Bosselut, Levy, Holtzman, Ennis, Fox, and
  Choi}]{bosselut2017simulating}
Antoine Bosselut, Omer Levy, Ari Holtzman, Corin Ennis, Dieter Fox, and Yejin
  Choi. 2018.
\newblock Simulating action dynamics with neural process networks.
\newblock In \emph{Proceedings of the 6th International Conference for Learning
  Representations (ICLR)}.

\bibitem[{Cho et~al.(2014)Cho, van Merri{\"e}nboer, Bahdanau, and
  Bengio}]{cho2014properties}
Kyunghyun Cho, Bart van Merri{\"e}nboer, Dzmitry Bahdanau, and Yoshua Bengio.
  2014.
\newblock On the properties of neural machine translation: Encoder{--}decoder
  approaches.
\newblock In \emph{Proceedings of {SSST}-8, Eighth Workshop on Syntax,
  Semantics and Structure in Statistical Translation}, pages 103--111, Doha,
  Qatar. Association for Computational Linguistics.

\bibitem[{Dalvi et~al.(2018)Dalvi, Huang, Tandon, Yih, and
  Clark}]{mishra2018tracking}
Bhavana Dalvi, Lifu Huang, Niket Tandon, Wen-tau Yih, and Peter Clark. 2018.
\newblock Tracking state changes in procedural text: a challenge dataset and
  models for process paragraph comprehension.
\newblock In \emph{Proceedings of the 2018 Conference of the North {A}merican
  Chapter of the Association for Computational Linguistics: Human Language
  Technologies, Volume 1 (Long Papers)}, pages 1595--1604, New Orleans,
  Louisiana. Association for Computational Linguistics.

\bibitem[{Dalvi et~al.(2019)Dalvi, Tandon, Bosselut, Yih, and
  Clark}]{dalvi2019everything}
Bhavana Dalvi, Niket Tandon, Antoine Bosselut, Wen-tau Yih, and Peter Clark.
  2019.
\newblock Everything happens for a reason: Discovering the purpose of actions
  in procedural text.
\newblock In \emph{Proceedings of the 2019 Conference on Empirical Methods in
  Natural Language Processing and the 9th International Joint Conference on
  Natural Language Processing (EMNLP-IJCNLP)}, pages 4496--4505, Hong Kong,
  China. Association for Computational Linguistics.

\bibitem[{Das et~al.(2018)Das, Munkhdalai, Yuan, Trischler, and
  McCallum}]{das2018building}
Rajarshi Das, Tsendsuren Munkhdalai, Xingdi Yuan, Adam Trischler, and Andrew
  McCallum. 2018.
\newblock Building dynamic knowledge graphs from text using machine reading
  comprehension.
\newblock In \emph{International Conference on Learning Representations}.

\bibitem[{Devlin et~al.(2019)Devlin, Chang, Lee, and
  Toutanova}]{devlin-etal-2019-bert}
Jacob Devlin, Ming-Wei Chang, Kenton Lee, and Kristina Toutanova. 2019.
\newblock {BERT}: Pre-training of deep bidirectional transformers for language
  understanding.
\newblock In \emph{Proceedings of the 2019 Conference of the North {A}merican
  Chapter of the Association for Computational Linguistics: Human Language
  Technologies, Volume 1 (Long and Short Papers)}, pages 4171--4186,
  Minneapolis, Minnesota. Association for Computational Linguistics.

\bibitem[{Dua et~al.(2019)Dua, Wang, Dasigi, Stanovsky, Singh, and
  Gardner}]{dua2019drop}
Dheeru Dua, Yizhong Wang, Pradeep Dasigi, Gabriel Stanovsky, Sameer Singh, and
  Matt Gardner. 2019.
\newblock {DROP}: A reading comprehension benchmark requiring discrete
  reasoning over paragraphs.
\newblock In \emph{Proceedings of the 2019 Conference of the North {A}merican
  Chapter of the Association for Computational Linguistics: Human Language
  Technologies, Volume 1 (Long and Short Papers)}, pages 2368--2378,
  Minneapolis, Minnesota. Association for Computational Linguistics.

\bibitem[{Gupta and Durrett(2019)}]{gupta2019tracking}
Aditya Gupta and Greg Durrett. 2019.
\newblock Tracking discrete and continuous entity state for process
  understanding.
\newblock In \emph{Proceedings of the Third Workshop on Structured Prediction
  for {NLP}}, pages 7--12, Minneapolis, Minnesota. Association for
  Computational Linguistics.

\bibitem[{Henaff et~al.(2017)Henaff, Weston, Szlam, Bordes, and
  LeCun}]{henaff2016tracking}
Mikael Henaff, Jason Weston, Arthur Szlam, Antoine Bordes, and Yann LeCun.
  2017.
\newblock Tracking the world state with recurrent entity networks.
\newblock In \emph{5th International Conference on Learning Representations,
  ICLR 2017}.

\bibitem[{Liu et~al.(2019)Liu, Ott, Goyal, Du, Joshi, Chen, Levy, Lewis,
  Zettlemoyer, and Stoyanov}]{liu2019roberta}
Yinhan Liu, Myle Ott, Naman Goyal, Jingfei Du, Mandar Joshi, Danqi Chen, Omer
  Levy, Mike Lewis, Luke Zettlemoyer, and Veselin Stoyanov. 2019.
\newblock Roberta: {A} robustly optimized {BERT} pretraining approach.
\newblock \emph{CoRR}, abs/1907.11692.

\bibitem[{Long et~al.(2016)Long, Pasupat, and Liang}]{long2016simpler}
Reginald Long, Panupong Pasupat, and Percy Liang. 2016.
\newblock Simpler context-dependent logical forms via model projections.
\newblock In \emph{Proceedings of the 54th Annual Meeting of the Association
  for Computational Linguistics (Volume 1: Long Papers)}, pages 1456--1465,
  Berlin, Germany. Association for Computational Linguistics.

\bibitem[{Paszke et~al.(2017)Paszke, Gross, Chintala, Chanan, Yang, DeVito,
  Lin, Desmaison, Antiga, and Lerer}]{paszke2017automatic}
Adam Paszke, Sam Gross, Soumith Chintala, Gregory Chanan, Edward Yang, Zachary
  DeVito, Zeming Lin, Alban Desmaison, Luca Antiga, and Adam Lerer. 2017.
\newblock Automatic differentiation in pytorch.

\bibitem[{Rajaby~Faghihi et~al.(2020)Rajaby~Faghihi, Mirzaee, Paliwal, and
  Kordjamshidi}]{faghihi2020latent}
Hossein Rajaby~Faghihi, Roshanak Mirzaee, Sudarshan Paliwal, and Parisa
  Kordjamshidi. 2020.
\newblock Latent alignment of procedural concepts in multimodal recipes.
\newblock In \emph{Proceedings of the First Workshop on Advances in Language
  and Vision Research}, pages 26--31.

\bibitem[{Rajpurkar et~al.(2016)Rajpurkar, Zhang, Lopyrev, and
  Liang}]{rajpurkar2016squad}
Pranav Rajpurkar, Jian Zhang, Konstantin Lopyrev, and Percy Liang. 2016.
\newblock {SQ}u{AD}: 100,000+ questions for machine comprehension of text.
\newblock In \emph{Proceedings of the 2016 Conference on Empirical Methods in
  Natural Language Processing}, pages 2383--2392, Austin, Texas. Association
  for Computational Linguistics.

\bibitem[{Santoro et~al.(2018)Santoro, Faulkner, Raposo, Rae, Chrzanowski,
  Weber, Wierstra, Vinyals, Pascanu, and Lillicrap}]{santoro2018relational}
Adam Santoro, Ryan Faulkner, David Raposo, Jack Rae, Mike Chrzanowski,
  Theophane Weber, Daan Wierstra, Oriol Vinyals, Razvan Pascanu, and Timothy
  Lillicrap. 2018.
\newblock Relational recurrent neural networks.
\newblock In \emph{Advances in neural information processing systems}, pages
  7299--7310.

\bibitem[{Schuler(2005)}]{schuler2005verbnet}
Karin~Kipper Schuler. 2005.
\newblock Verbnet: A broad-coverage, comprehensive verb lexicon.

\bibitem[{Seo et~al.(2017)Seo, Min, Farhadi, and Hajishirzi}]{seo2016query}
Min~Joon Seo, Sewon Min, Ali Farhadi, and Hannaneh Hajishirzi. 2017.
\newblock Query-reduction networks for question answering.
\newblock In \emph{5th International Conference on Learning Representations,
  {ICLR} 2017, Toulon, France, April 24-26, 2017, Conference Track
  Proceedings}. OpenReview.net.

\bibitem[{Sukhbaatar et~al.(2015)Sukhbaatar, szlam, Weston, and
  Fergus}]{weston2014memory}
Sainbayar Sukhbaatar, arthur szlam, Jason Weston, and Rob Fergus. 2015.
\newblock \href
  {https://proceedings.neurips.cc/paper/2015/file/8fb21ee7a2207526da55a679f0332de2-Paper.pdf}
  {End-to-end memory networks}.
\newblock In \emph{Advances in Neural Information Processing Systems},
  volume~28, pages 2440--2448. Curran Associates, Inc.

\bibitem[{Tandon et~al.(2018)Tandon, Dalvi, Grus, Yih, Bosselut, and
  Clark}]{tandon2018reasoning}
Niket Tandon, Bhavana Dalvi, Joel Grus, Wen-tau Yih, Antoine Bosselut, and
  Peter Clark. 2018.
\newblock Reasoning about actions and state changes by injecting commonsense
  knowledge.
\newblock In \emph{Proceedings of the 2018 Conference on Empirical Methods in
  Natural Language Processing}, pages 57--66, Brussels, Belgium. Association
  for Computational Linguistics.

\bibitem[{Weston et~al.(2015)Weston, Bordes, Chopra, Rush, van Merriënboer,
  Joulin, and Mikolov}]{weston2015towards}
Jason Weston, Antoine Bordes, Sumit Chopra, Alexander~M. Rush, Bart van
  Merriënboer, Armand Joulin, and Tomas Mikolov. 2015.
\newblock Towards ai-complete question answering: A set of prerequisite toy
  tasks.
\newblock Cite arxiv:1502.05698.

\bibitem[{Wolf et~al.(2019)Wolf, Debut, Sanh, Chaumond, Delangue, Moi, Cistac,
  Rault, Louf, Funtowicz, Davison, Shleifer, von Platen, Ma, Jernite, Plu, Xu,
  Scao, Gugger, Drame, Lhoest, and Rush}]{Wolf2019HuggingFacesTS}
Thomas Wolf, Lysandre Debut, Victor Sanh, Julien Chaumond, Clement Delangue,
  Anthony Moi, Pierric Cistac, Tim Rault, Rémi Louf, Morgan Funtowicz, Joe
  Davison, Sam Shleifer, Patrick von Platen, Clara Ma, Yacine Jernite, Julien
  Plu, Canwen Xu, Teven~Le Scao, Sylvain Gugger, Mariama Drame, Quentin Lhoest,
  and Alexander~M. Rush. 2019.
\newblock Huggingface's transformers: State-of-the-art natural language
  processing.
\newblock \emph{ArXiv}, abs/1910.03771.

\bibitem[{Yagcioglu et~al.(2018)Yagcioglu, Erdem, Erdem, and
  Ikizler-Cinbis}]{yagcioglu2018recipeqa}
Semih Yagcioglu, Aykut Erdem, Erkut Erdem, and Nazli Ikizler-Cinbis. 2018.
\newblock Recipeqa: A challenge dataset for multimodal comprehension of cooking
  recipes.
\newblock In \emph{EMNLP}.

\bibitem[{Yang et~al.(2019)Yang, Dai, Yang, Carbonell, Salakhutdinov, and
  Le}]{yang2019xlnet}
Zhilin Yang, Zihang Dai, Yiming Yang, Jaime Carbonell, Russ~R Salakhutdinov,
  and Quoc~V Le. 2019.
\newblock Xlnet: Generalized autoregressive pretraining for language
  understanding.
\newblock In \emph{Advances in neural information processing systems}, pages
  5753--5763.

\bibitem[{Yang et~al.(2018)Yang, Qi, Zhang, Bengio, Cohen, Salakhutdinov, and
  Manning}]{yang2018hotpotqa}
Zhilin Yang, Peng Qi, Saizheng Zhang, Yoshua Bengio, William Cohen, Ruslan
  Salakhutdinov, and Christopher~D. Manning. 2018.
\newblock {H}otpot{QA}: A dataset for diverse, explainable multi-hop question
  answering.
\newblock In \emph{Proceedings of the 2018 Conference on Empirical Methods in
  Natural Language Processing}, pages 2369--2380, Brussels, Belgium.
  Association for Computational Linguistics.

\bibitem[{Zhang et~al.(2021)Zhang, Geng, Qin, Wu, and
  Jiang}]{zhang2020knowledge}
Zhihan Zhang, Xiubo Geng, Tao Qin, Yunfang Wu, and Daxin Jiang. 2021.
\newblock Knowledge-aware procedural text understanding with multi-stage
  training.
\newblock In \emph{Proceedings of the Web Conference 2021}.

\end{thebibliography}
\bibliographystyle{acl_natbib}

% \appendix

% \section{Example Appendix}
% \label{sec:appendix}

% This is an appendix.

\end{document}